\documentclass{bmvc2k}

\title{Privacy-preserving datasets by capturing feature distributions with Conditional VAEs}

\addauthor{Francesco Di Salvo}{francesco.di-salvo@uni-bamberg.de}{1}
\addauthor{David Tafler}{david-elias.tafler@stud.uni-bamberg.de}{1}
\addauthor{Sebastian Doerrich}{sebastian.doerrich@uni-bamberg.de}{1}
\addauthor{Christian Ledig}{christian.ledig@uni-bamberg.de}{1}

\addinstitution{
 xAILab Bamberg\\
 University of Bamberg \\
 Bamberg, Germany
}

\runninghead{Di Salvo et al.}{Capturing feature distributions with CVAEs}

\def\eg{\emph{e.g}\bmvaOneDot}

\usepackage{booktabs}
\usepackage{graphicx}
\usepackage{svg}
\usepackage{xcolor,colortbl}

\definecolor{grey}{HTML}{f2f2f2}
\definecolor{pastelgreen}{HTML}{81dca2}
\definecolor{pastelorange}{HTML}{f9b48b}

\usepackage{comment}

\usepackage{algorithm}
\usepackage[noend]{algorithmic}
\UseRawInputEncoding

\usepackage{booktabs}
\usepackage{multirow}
\begin{document}

\maketitle

\begin{abstract}

Large and well-annotated datasets are essential for advancing deep learning applications, however often costly or impossible to obtain by a single entity. In many areas, including the medical domain, approaches relying on data sharing have become critical to address those challenges. While effective in increasing dataset size and diversity, data sharing raises significant privacy concerns. Commonly employed anonymization methods based on the $k$-anonymity paradigm often fail to preserve data diversity, affecting model robustness. This work introduces a novel approach using Conditional Variational Autoencoders ($\texttt{CVAE}$s) trained on feature vectors extracted from large pre-trained vision foundation models. Foundation models effectively detect and represent complex patterns across diverse domains, allowing the $\texttt{CVAE}$ to faithfully capture the embedding space of a given data distribution to generate (sample) a diverse, privacy-respecting, and potentially unbounded set of synthetic feature vectors. Our method notably outperforms traditional approaches in both medical and natural image domains, exhibiting greater dataset diversity and higher robustness against perturbations while preserving sample privacy. These results underscore the potential of generative models to significantly impact deep learning applications in data-scarce and privacy-sensitive environments. The source code is available at \href{https://github.com/francescodisalvo05/cvae-anonymization}{github.com/francescodisalvo05/cvae-anonymization}.
\end{abstract}

\begin{figure}[htbp!]
\begin{center}
\includegraphics[width=0.9\linewidth]{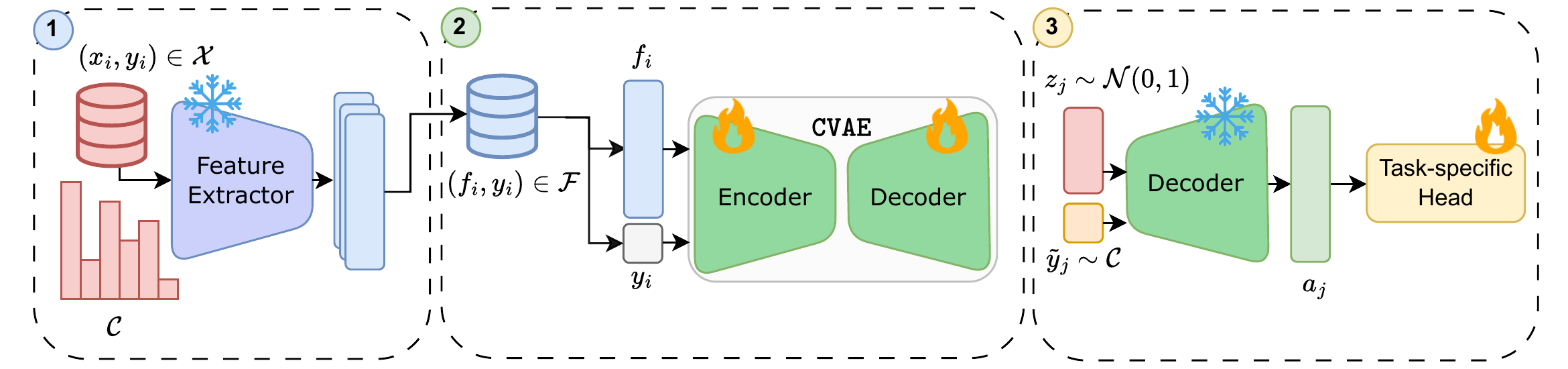}
\end{center}
   \caption{\label{pull_figure}Illustration of our proposed anonymization approach. Given an image dataset $(x_i,y_i) \in \mathcal{X}$ with categorical class distribution $\mathcal{C}$, we first utilize a large pre-trained model to extract and store feature embeddings and corresponding labels $(f_i,y_i) \in \mathcal{F}$. These embeddings capture both local and contextual information while inherently reducing dimensionality. Subsequently, the embeddings are used during training of a Conditional Variational Autoencoder (\texttt{CVAE}) to capture the training distribution conditioned on the respective class labels $y_i$. Finally, we train a task-specific head 
  while dynamically generating new synthetic feature vectors $a_j$ conditioned on class labels $\tilde y_j \sim C$ through \texttt{CVAE}'s frozen decoder. This not only ensures data anonymity but also increases data diversity and model robustness.}
\end{figure}
\section{Introduction}

Over the past decade, Deep Neural Networks have made significant progress, impacting a wide range of industries. A key driver of this progress has been the increasing availability of large, well-annotated datasets. However, such datasets are rare in several domains, such as medical image analysis \cite{altaf2019going}. While data-sharing between institutions offers a potential solution, it also raises privacy concerns over \textit{personal data}. To this extent, data privacy has been a focal point for decades. \newline 

For instance, the $k$-\textit{anonymity} paradigm \cite{sweeney2002k} was introduced in 2002 to obscure user data in relational databases, ensuring that no individual could be distinguished from at least $k{-}1$ others. Building on this idea, the $k$-Same \cite{newton2005preserving} method emerged to address privacy concerns in surveillance technologies. It aggregates disjoint clusters of $k$ data points (\textit{i.e.}, faces) with their centroid, either in the pixel space or in the eigenspace.

With the advent of generative adversarial networks (GANs) \cite{goodfellow2020generative,creswell2018generative}, new possibilities for generating synthetic images have emerged. However, there is a risk of embedding personal information in the latent space. To address this, methods such as $k$-Same-Net \cite{meden2018k} enforced $k$-anonymity within GANs' latent space during training. Furthermore, GANs have struggled with issues such as mode collapse, where models fail to produce diverse outputs, thereby limiting their generalizability \cite{saxena2021generative,9207181}. Orthogonally, recent advancements have shifted focus to vision foundation models \cite{radford2021learning,caron2021emerging,oquab2023dinov2}, typically trained in a self-supervised fashion on large unlabeled datasets. These large models using vision transformers \cite{dosovitskiy2020image} excel at capturing high-level complex patterns and contextual information across different domains. Moreover, their robustness against image corruptions and distribution shifts \cite{paul2022vision} is an established benefit. Researchers are now proposing foundation models for specific domains including healthcare \cite{tu2024towards,zhou2023foundation,li2024llava}, climate \cite{nguyen2023climax}, geospatial data \cite{mai2022towards}, and more. The rich and low-dimensional feature representation derived from these models makes them a suitable alternative (to the higher dimensional raw images) for pursuing downstream tasks, such as classification.  

Building on the potential of those highly complex and informative representations, as illustrated in Figure \ref{pull_figure}, we train a Conditional Variational Autoencoder (\texttt{CVAE}) on the feature space of a pre-trained foundation model to accurately capture the training distribution. The generative process, conditioned on class labels, effectively mimics the original distribution while enhancing privacy and expanding the diversity of the generated feature vectors. After capturing the training data distribution with our \texttt{CVAE}, we demonstrate a paradigm that no longer relies on disclosing the original images (or their embeddings) but simply the frozen \texttt{CVAE} decoder allowing the sampling of any required dataset size. Despite tremendous benefits with respect to data and patient privacy, this also substantially reduces the required data exchange from an order of gigabytes (images) to a few megabytes (decoder weights). \newline

In summary, our contributions are:

\begin{itemize}
	\item We propose a novel anonymization paradigm using Conditional Variational Autoencoders (\texttt{CVAE}s) trained on the feature space of foundation models, which provides a more informative and compact representation than the traditional image space.
	\item We demonstrate superior feature diversity and classification performance with our \texttt{CVAE} compared to reference approaches across both medical and natural datasets. This also highlights the effectiveness of foundation model's feature representation, even in areas beyond its initial training focus. 
	\item We show that the diverse feature representation obtained via dynamic sampling and decoding through \texttt{CVAE}'s decoder enhances model robustness against perturbations.
\end{itemize}

\section{Related works}

As digital data usage expands, the need for robust privacy measures becomes critical. The $k$-anonymity framework, proposed by Sweeney \cite{sweeney2002k}, addresses this by ensuring that an individual cannot be distinguished from at least $k-1$ others within database entries. This method involves generalizing or suppressing specific identifying attributes such as zip code, gender, and other sensitive information. While effective at masking identities, this approach often leads to significant data loss, reducing the utility of the data. With the rise of video surveillance systems in New York, a seminal work \cite{newton2005preserving} introduced $k$-Same for face de-identification, translating these principles to more complex scenarios, including the \textit{pixel space} and the \textit{eigenspace}. This is achieved by collapsing groups of $k$ elements into a single representative point, namely the centroid. Despite its application in real-world medical systems \cite{gkoulalas2014publishing}, the inherited data loss and lower data diversity can substantially affect the generalization \cite{8758818,yu-etal-2022-data} and robustness \cite{gowal2021improving,larson-etal-2020-iterative} of machine learning based solutions, particularly in domains where data is already scarce.

Moreover, building on the capabilities of GANs to generate high-quality synthetic samples, $k$-Same-Net \cite{meden2018k} applies the $k$-anonymity paradigm within GANs' latent space for face de-identification. Enhancing this further, $k$-SALSA \cite{jeon2022k} recently introduced local style alignment to preserve granular details in retinal images, while PLAN \cite{pennisi2023privacy} introduces an auxiliary identity classifier to prevent sample collisions. However, GANs still face significant challenges due to their high computational demands, training instability, and mode collapse \cite{9207181, saxena2021generative}, which can affect their efficiency and reliability.

Before the advent of GANs, Variational Autoencoders (VAEs) have already demonstrated their generative capabilities. For face de-identification, Yang et al. \cite{yang2022cluster} generate realistic yet untraceable images by clustering and re-synthesizing facial features. Taehoon et al. \cite{8744221} enforce privacy by introducing noise into the latent vectors, obscuring identities before reconstruction. Other than the image domain, VAEs have been employed for anonymizing speaker data \cite{perero2022x} and tabular data \cite{nguyen2023climax}, as well. Conditional Variational Autoencoders (CVAEs) enhance the versatility of VAEs by allowing the generative process to be conditioned on additional labels or attributes, effectively tailoring the output to specific data characteristics. Hajihassnai et al. \cite{hajihassnai2021obscurenet} introduce ObscureNet \cite{hajihassnai2021obscurenet}, a CVAE-like architecture to anonymize sensor data, that conditions the latent space on private attributes (\eg, age, gender). The architecture consists of one input encoder, \textit{multiple external discriminators} (one per attribute), and a conditional decoder. Apart from anonymization, the conditional generative process of CVAEs has been used to augment training data \cite{pesteie2019adaptive,fajardo2021oversampling}, addressing issues such as data scarcity and imbalance. 

In contrast, our approach first enhances the conditional generative process by shifting the generation from the pixel space to the \textit{embedding space}, utilizing large pre-trained foundation models. This transition enables a more efficient feature generation, thanks to the more compact and informative feature representation. In contrast to ObscureNet \cite{hajihassnai2021obscurenet}, we employ one conditional encoder rather than multiple discriminators, which allows our architecture to scale more efficiently with the number of classes. Furthermore, instead of simply augmenting our training dataset, we focus on \textit{mimicking} its embedding-based training distribution, aiming to replicate a diverse and privacy-aware distribution. We further integrate this generative process during training of downstream tasks, which allows us to dynamically generate new samples until convergence. 

\section{Method}

The proposed approach to capture and anonymize a given training dataset is illustrated in Figure \ref{pull_figure}. Initially, we describe the process of feature extraction through vision foundation models. This is followed by a description of \texttt{CVAE}'s training procedure. We then introduce two anonymization strategies: the first generates a persistent synthetic replica of the training dataset, while the second continuously generates data on the fly for downstream tasks. 

\subsection{Feature extraction} 

We employ a large pre-trained model to extract and subsequently store feature embeddings and their associated labels $(f_i,y_i) \in \mathcal{F}$ for each labeled input image $(x_i,y_i) \in \mathcal{X}$, with an associated categorical class distribution $\mathcal{C}$. The extracted feature representations $f_i$ represent the positional information of $x_i$ within the feature space of the selected feature extractor. Here, close points are more likely to share similar image patterns while distant points indicate semantic differences, potentially across different classes. This descriptive property of the feature space is beneficial for effectively capturing the training distribution. It is much less pronounced in the original pixel space, which suffers from inherent information redundancy and the challenge of calculating meaningful distances in high-dimensional spaces. Additionally, from a storage perspective, while a conventional 2D RGB image of dimensions $224{\times}224$ occupies approximately $150{,}000$ pixels, the corresponding image embeddings typically have around $1{,}000$ components, reducing data storage by several orders of magnitude. For our experiments, we rely on DINOv2 ViT-B/14 \cite{oquab2023dinov2}, one of the most recent open-source foundation models that achieved state-of-the-art performance in a variety of tasks, demonstrating the high quality of their latent feature representations.

\subsection{\texttt{CVAE} training} 

Conditional Variational Autoencoders extend traditional Variational Autoencoders by integrating specific attributes, such as class labels, into the data generation process. This extension allows the \texttt{CVAE} to not only produce data that closely resembles the original samples in terms of content but also aligns with the conditioned attributes, thereby improving both the relevance and specificity of the generated data. In our setup, the \texttt{CVAE} is trained to accurately learn and replicate the distribution of a feature dataset $\mathcal{F}$, conditioned on the corresponding labels, i.e., $(f_i,y_i=c)$. This training allows us to subsequently generate new \textit{anonymous feature vectors} $a_j$ for a chosen class $\tilde y_j$ that reflect the original training distribution. We can thus utilize only the pre-trained decoder to sample arbitrarily many samples per class $c$, adhering to the original categorical class distribution $\mathcal{C}$. This approach ensures that the generated data maintains consistency with the original dataset while enhancing privacy and utility, likewise.

\subsection{Offline anonymization} 

The first way to anonymize our dataset is via storing a persistent anonymized replica of the data. To this end, once the \texttt{CVAE} is appropriately trained, we generate synthetic feature vectors that reflect the initial data size and class distribution. This anonymization process is described in Algorithm \ref{alg:anonymize} and requires a pre-trained \texttt{CVAE} model, a number of synthetic samples $N$ and a categorical distribution $\mathcal{C}$ representing the class probabilities of the original dataset. This ensures the generation of similar class proportions in the synthetic dataset. For each of the desired $N$ data points (step 2), a class label $\tilde y_j$ is sampled from $\mathcal{C}$ (step 3). Next, a latent variable $z_j$ is drawn from a standard normal distribution (step 4) to serve as a stochastic input for data generation. These sampled variables $(z_j,\tilde y_j)$ are fed into \texttt{CVAE}'s decoder, which generates a new data point $a_j$ conditioned on the class label (step 5). Finally, both the generated data point and its label are then stored (step 6). Upon completion, the algorithm outputs the anonymized datasets, consisting of these synthetically generated feature vectors and their labels, effectively balancing privacy preservation with the utility of the dataset for downstream tasks.

\begin{algorithm}
\caption{\texttt{anonymize}($\texttt{CVAE},\mathcal{C},N$)
}\label{alg:anonymize}
\begin{algorithmic}[1]
	\STATE $\mathcal{A} \gets [ \, ] $ \hfill \COMMENT{initialize anonymized dataset}
	\FOR{$j = 1, 2, \ldots, N$}  
		\STATE $\tilde y_j \sim \mathcal{C}$ \hfill \COMMENT{sample $\tilde y_j$ from class categorical distribution $\mathcal{C}$}
		\STATE $z_j \sim \mathcal{N}(0,1)$ \hfill \COMMENT{standard normal sampling}
		\STATE $a_j \gets \texttt{CVAE.decoder}(z_j,\tilde y_j)$ \hfill \COMMENT{conditional decoding}
		\STATE $\mathcal{A}\texttt{.append}((a_j,\tilde y_j))$ \hfill \COMMENT{append the generated tuple}
	\ENDFOR
	
\RETURN $\mathcal{A}$
\end{algorithmic}
\end{algorithm}

\subsection{Online anonymization}
\label{sec:data_free}

Another way to ensure data anonymization is to eliminate the use of persistent datasets entirely. Our proposed \texttt{CVAE} model exemplifies this approach by replacing the traditional dataset with the \texttt{CVAE}'s pre-trained decoder. Specifically, we iteratively generate new data on the fly during the training of our task-specific head, meaning no persistent data is required anymore. This method not only overcomes the need to store or send large volumes of sensitive data but also enables the sharing of a simple model (\texttt{CVAE}'s decoder) capable of replicating the training data distribution at an arbitrary incidence rate without actual data exchange. Moreover, in the context of model sharing, such as in federated learning \cite{zhang2021survey,rodriguez2023survey}, our approach offers additional security benefits. While federated learning allows for task-specific models trained on private data to be shared without exchanging the data itself, it is not without risks; model weights can potentially reveal training data characteristics \cite{nasr2019comprehensive,app12199901}. \newline

Our approach iteratively generates anonymized and stochastic data. This can introduce an additional layer of security, further mitigating the risks associated with traditional data- and model-sharing approaches.

\section{Experimental results}

Initially, we outline the process of feature extraction and the training of our Conditional Variational Autoencoder (\texttt{CVAE}). Subsequently, we evaluate the classification performance and feature diversity using our anonymized method compared to $k$-Same \cite{newton2005preserving}, the foundational anonymization method that influenced machine-learning-related applications, such as GANs \cite{meden2018k}. Additionally, we evaluate the classification performance under inference perturbations, showcasing the robustness of the generated feature representation. Lastly, we perform a qualitative analysis of the initial feature space and the anonymized one, in order to visually comprehend the representativeness of the proposed method. 
\subsection{Feature extraction and \texttt{CVAE} training}

For feature extraction, we used the DINOv2 ViT-B/14 foundation model \cite{oquab2023dinov2}, which has been trained on 142 million natural images and outputs image embeddings of size $768$. To optimize the training process of our \texttt{CVAE}, these embeddings were pre-generated and stored on disk \cite{nakata2022revisiting,doerricharchut2024kNNIntegration}. For datasets lacking official validation splits, we extracted a stratified sample $(10\%)$ from the training data to ensure representativeness. Note that both validation and test sets were kept non-anonymized, preserving their semantic integrity.  

The \texttt{CVAE} architecture comprises a conditional encoder and a conditional decoder. The encoder first concatenates input features with one-hot class labels and then processes it through two linear layers having 256 and 100 dimensions, respectively. The latent variable $z$ is again concatenated with the one-hot class labels and passed to a symmetric decoder that mirrors the encoder's architecture. We train the \texttt{CVAE} with the Adam optimizer, learning rate of $0.001$, and early stopping. In addition to the MSE loss, a KL-divergence loss is employed to enforce a standard normal prior in the latent space weighted with $\beta = 0.1$, prioritizing the reconstruction. These hyperparameters are standardized across all experiments to ensure consistent comparisons.

\subsection{Feature diversity and downstream performance}
\label{sec:downstream}

\textbf{Evaluation metrics} \, The $k$-Same method simplifies anonymization by collapsing groups of $k$ data points into their centroid. While effective for anonymization, this approach often results in significant information loss and increased data sparsity. Conversely, \texttt{CVAE} effectively maintains the initial distribution's representation, enabling the generation of more samples than originally used, without the compromise on diversity. Yu et al. \cite{yu-etal-2022-data} estimate data diversity through the \textit{convex hull} and the \textit{maximum dispersion}, defined as the sum of all pairwise distances within the (anonymized) dataset. Although calculating the convex hull is impractical for datasets with fewer samples than dimensions, it is trivial to demonstrate that \texttt{CVAE} achieves higher maximum dispersion compared to $k$-Same, as the latter's regression to the centroid inherently reduces pairwise distances. \newline

To provide a more nuanced comparison, we analyze the \textit{average nearest neighbor distance} $\mathcal{D}$ between the original feature set $\mathcal{F}$ and the anonymized feature set $\mathcal{A}$ (both of size $N$). This is formally defined in Equation \ref{eq_distance}:

\begin{equation}
	\label{eq_distance}
	\mathcal{D}(\mathcal{A},\mathcal{F}) = \frac {1} {N} \sum_{j = 1}^N d_{\text{min}}(a_j,\mathcal{F})
\end{equation}

where $d_{\text{min}}(a_i,\mathcal{F})$ is the minimum Euclidean distance between $a_j \in \mathcal{A}$ and all $f_i \in \mathcal{F}$. Intuitively, this is closely related to the \textit{mean of the minimum LPIPS} \cite{zhang2018perceptual} used in PLAN \cite{pennisi2023privacy}. In fact, the latter measures the average minimum distance between generated samples and their closest real ones, using the activations of a pre-trained network. Moreover, an equally important goal of any anonymization technique is to preserve downstream performance while ensuring data privacy. In our evaluation, considering image classification as the downstream task, we use the \textit{area under the receiver operating curve} (AUC) as the evaluation metric. This allows us to take into account different class imbalance ratios across the evaluated datasets. Following prior works in feature space evaluation \cite{oquab2023dinov2, caron2021emerging}, we train a linear layer on top of the anonymized embeddings using the Adam optimizer with a learning rate of $0.001$ and early stopping. To fairly compare our method against $k$-Same, we replicate the \textit{same exact class proportions} of the original dataset. \newline 

\noindent \textbf{Reference method and datasets} \, Consistent with prior works \cite{jeon2022k,pennisi2023privacy}, we evaluate the performance of our \texttt{CVAE} against $k$-Same using $k \in \{2,5,10,15\}$. We exclude the previously mentioned GAN-based methods such as $k$-SALSA \cite{jeon2022k} and PLAN \cite{pennisi2023privacy}, since they applied $k$-Same within GAN's latent space to generate anonymous synthetic \textit{images}. Our analysis begins with a wide range of real-world, small-sized, medical datasets to underscore the clinical relevance of our method. From the MedMNIST+ collection \cite{yang2023medmnist,doerrich2024rethinking}, we use BreastMNIST with resolution $224{\times}224$ (breast ultrasound, binary, 780 samples). Additionally, we include datasets from the MedIMeta collection \cite{woerner2024comprehensive}: Skin Lesion (dermatoscopy, multi-class, 1011 samples) and Axial Organ slices (CT, multi-class, 1645 samples). Another challenging dataset is OCTDL \cite{daneshjou2022disparities} (OCT imaging, multi-class, 2064 samples), which exhibits a severe class imbalance. To extend our evaluation to the natural image domain, we selected multi-class datasets from the \texttt{torchvision.datasets} module \footnote{\href{https://pytorch.org/vision/stable/datasets.html}{https://pytorch.org/vision/stable/datasets.html}}, focusing on those with moderate sample sizes. This includes STL-10 \cite{coates2011analysis}, DTD \cite{cimpoi14describing}, Oxford-Pets \cite{parkhi2012cats}, and FGVC-Aircraft \cite{maji13fine-grained}, allowing us to cover a wide spectrum of general imaging tasks. \newline

\begin{figure}[htbp!]
\begin{center}
\includegraphics[width=0.9\linewidth]{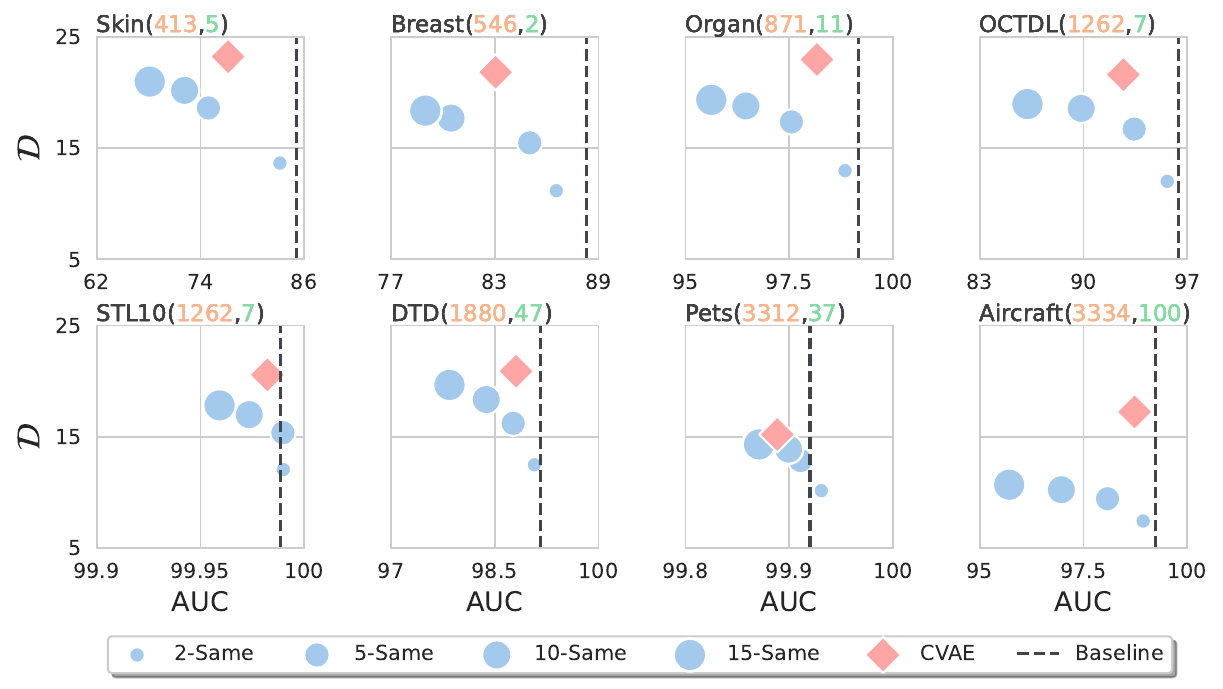}
\end{center}
   \caption{\label{fig:scatter} Classification performance (AUC) and average nearest neighbor distance ($\mathcal{D}$) on medical (top row) and non-medical (bottom row) datasets. We report in brackets the number of ({\color{pastelorange} training samples},{\color{pastelgreen}classes}). Our objective is to maximize both metrics (top-right corner). The vertical line represents the \textit{baseline} performance achieved without anonymization.}
 \end{figure}

\noindent \textbf{Results} \, Figure \ref{fig:scatter} presents a scatterplot with AUC on the $x$-axis and average nearest neighbor distance $\mathcal{D}$ on the $y$-axis, positioning the most effective methods towards the top-right corner. To ensure the reliability of our findings, all results are averaged across three different seed runs.  \texttt{CVAE} consistently exhibits both high diversity and high downstream performance ${(\mathcal{D} \uparrow , \text{AUC} \uparrow)}$. In contrast, as expected, lower $k$-Same configurations $(k=2,5)$ achieve comparable downstream performance but lack diversity ${(\mathcal{D} \downarrow , \text{AUC} \uparrow)}$, whereas higher $k$-Same settings $(k=10,15)$ increase diversity at the expense of performance ${(\mathcal{D} \uparrow , \text{AUC} \downarrow)}$. Interestingly, consistent results are observed with the OCTDL dataset, which is particularly challenging. Specifically, it presents a severe class imbalance, with a majority class of 752 samples and a minority of just 13. Furthermore, the medical results hold true on natural datasets as well, although the differences in classification performance are less pronounced, partly because these datasets were included in the DINOv2 training set \cite{oquab2023dinov2}. 
 
\subsection{Robustness of adaptive data sampling} 

This experiment evaluates the robustness of our proposed \texttt{CVAE} method, which dynamically generates new samples $a_j$ of class $\tilde y_j \sim \mathcal{C}$ at every batch. Here, the inherent data diversity provided by our method is expected to contribute positively to model robustness. To evaluate this claim, a common approach involves testing against image corruptions that simulate real-world distortions \cite{hendrycks2019benchmarking}. However, these corruptions in the pixel space might already be mitigated by the inherent robustness of foundation models \cite{paul2022vision}. Therefore, we propose a perturbation test by injecting Gaussian noise into the test feature embeddings. We systematically apply this noise with zero mean and a gradually increasing standard deviation $\sigma = \{1, 2, 3\}$. Here, we evaluate the performance of \texttt{CVAE} against $k$-Same. Therefore, we first train our \texttt{CVAE} and then use its decoder to continuously generate new samples during training. Instead of only sampling from the latent space with a standard Gaussian distribution $(\mu=0, \sigma^2=1)$ as reported in step 4 of Algorithm \ref{alg:anonymize}, we sample with variance $\sigma^2 = \{0.5,1.0,1.5\}$. This allows us to vary the prototypicality of the generated vectors. Specifically, since we enforce a standard normal distribution $(\sigma^2=1)$ during training, a lower sampling variance during data generation yields samples closer to the class prototypes, and vice versa. The classification pipeline follows the previous settings, reporting the average AUC over three seed runs. \newline

\noindent \textbf{Results} \, Table \ref{robustness_table} shows the AUC observed on the clean test feature vectors $(\sigma=0)$ and on those subjected to varying levels of Gaussian noise. \texttt{CVAE} consistently ranks among the top performers. Although \texttt{CVAE} may initially show slightly lower performance in a noise-free environment, it substantially outperforms $k$-Same as the noise level increases. This is particularly pronounced in the Skin Lesion and OCTDL, where \texttt{CVAE} $(\sigma^2=0.5)$ outperforms $2$-Same at $\sigma=3$ by 4.2\% and 3.2\%, respectively. 
Notably, this lower sampling variance remarkably improves performance and robustness across all datasets, suggesting that a higher prototypicality might be beneficial in settings with limited and imbalanced data. These results underscore the ability of our approach to enhance model robustness across diverse domains, each presenting unique challenges related to data size and data imbalance.

\begin{table}[!h]
\centering
\setlength{\tabcolsep}{3.5pt}
\scriptsize
\begin{tabular}{c ccccccccccccccccc}
\toprule
	
	& \multicolumn{15}{c}{\textbf{AUC} $\uparrow$} \\ \cmidrule(lr){2-17}
	
	& \multicolumn{4}{c}{\textbf{Skin}}
	& \multicolumn{4}{c}{\textbf{Breast}}
	& \multicolumn{4}{c}{\textbf{Organ}}
	& \multicolumn{4}{c}{\textbf{OCTDL}} \\ 
		\cmidrule(r){2-5} \cmidrule(r){6-9} \cmidrule(r){10-13} \cmidrule(r){14-17}



\textbf{$\sigma$}
	& \cellcolor{grey} 0 & 1 & 2 & 3
	& \cellcolor{grey} 0 & 1 & 2 & 3
	& \cellcolor{grey} 0 & 1 & 2 & 3
	& \cellcolor{grey} 0 & 1 & 2 & 3 \\ \midrule

$2$-Same 
	& \cellcolor{grey} \textbf{83.2} 
	& \textbf{78.9} 
	& \textbf{72.3} 
	& \textbf{67.2} 
	& \cellcolor{grey} \textbf{86.6} 
	& \textbf{83.5} 
	& 77.2 
	& 72.4 
	& \cellcolor{grey} \textbf{98.8} 
	& \textbf{98.2} 
	& \textbf{96.0} 
	& \textbf{92.2} 
	& \cellcolor{grey} \textbf{95.6} 
	& \textbf{92.3} 
	& \textbf{84.7} 
	& 77.9 
	\\ 
$5$-Same 
	& \cellcolor{grey} 74.9 
	& 71.7 
	& 66.9 
	& 63.1 
	& \cellcolor{grey} \textbf{85.0} 
	& \textbf{82.3} 
	& 76.6 
	& 72.2 
	& \cellcolor{grey} 97.6 
	& 96.9 
	& 94.8 
	& \textbf{91.2} 
	& \cellcolor{grey} \textbf{93.4} 
	& 90.0 
	& 82.2 
	& 75.4 
	\\ 
$10$-Same 
	& \cellcolor{grey} 72.2 
	& 69.8 
	& 65.6 
	& 62.2 
	& \cellcolor{grey} 80.5 
	& 78.4 
	& 72.8 
	& 68.7 
	& \cellcolor{grey} 96.5 
	& 95.6 
	& 93.3 
	& 89.6 
	& \cellcolor{grey} 89.8 
	& 87.8 
	& 82.6 
	& 76.8 
	\\ 
$15$-Same 
	& \cellcolor{grey} 68.1 
	& 66.5 
	& 63.5 
	& 60.7 
	& \cellcolor{grey} 79.0 
	& 76.7 
	& 71.8 
	& 67.9 
	& \cellcolor{grey} 95.6 
	& 94.7 
	& 92.1 
	& 88.3 
	& \cellcolor{grey} 86.2 
	& 84.5 
	& 79.4 
	& 74.5 
	\\ \midrule
$\texttt{CVAE} | \tiny{\sigma^2=0.5}$ 
	& \cellcolor{grey} \textbf{82.4} 
	& \textbf{80.5} 
	& \textbf{75.9} 
	& \textbf{71.4} 
	& \cellcolor{grey} 82.9 
	& 81.2 
	& \textbf{77.7} 
	& \textbf{74.2} 
	& \cellcolor{grey} 98.2 
	& 97.6 
	& \textbf{95.3} 
	& 91.1 
	& \cellcolor{grey} 93.3 
	& \textbf{91.2} 
	& \textbf{86.5} 
	& \textbf{81.1} 
	\\ 
$\texttt{CVAE} | \tiny{\sigma^2=1.0}$ 
	& \cellcolor{grey} 78.9 
	& 75.4 
	& 69.9 
	& 65.4 
	& \cellcolor{grey} 82.6 
	& 81.2 
	& \textbf{77.8} 
	& \textbf{73.8} 
	& \cellcolor{grey} 98.5 
	& 97.8 
	& 95.2 
	& 90.5 
	& \cellcolor{grey} 92.6 
	& 90.2 
	& 84.5 
	& \textbf{78.4} 
	\\ 
$\texttt{CVAE} | \tiny{\sigma^2=1.5}$ 
	& \cellcolor{grey} 73.2 
	& 69.1 
	& 63.9 
	& 60.3 
	& \cellcolor{grey} 83.5 
	& 81.9 
	& 76.9 
	& 72.0 
	& \cellcolor{grey} \textbf{98.6} 
	& \textbf{98.0} 
	& \textbf{95.3} 
	& 90.4 
	& \cellcolor{grey} 91.0 
	& 88.2 
	& 82.1 
	& 75.7 
	\\ 

	\bottomrule


\end{tabular}

\caption{\label{robustness_table} Area Under the Receiver Operating Curve (AUC $\uparrow$) calculated on the clean test embeddings ($\sigma = 0$) and across three replicas subject to Gaussian noise with zero mean and standard deviation $\sigma = \{1,2,3\}$. The top two results are displayed in \textbf{bold}.}
\end{table}
\subsection{Qualitative results}
\label{sec:qualitative}

\begin{figure}[h]
	\centering     
	\includegraphics[width=0.80\linewidth]{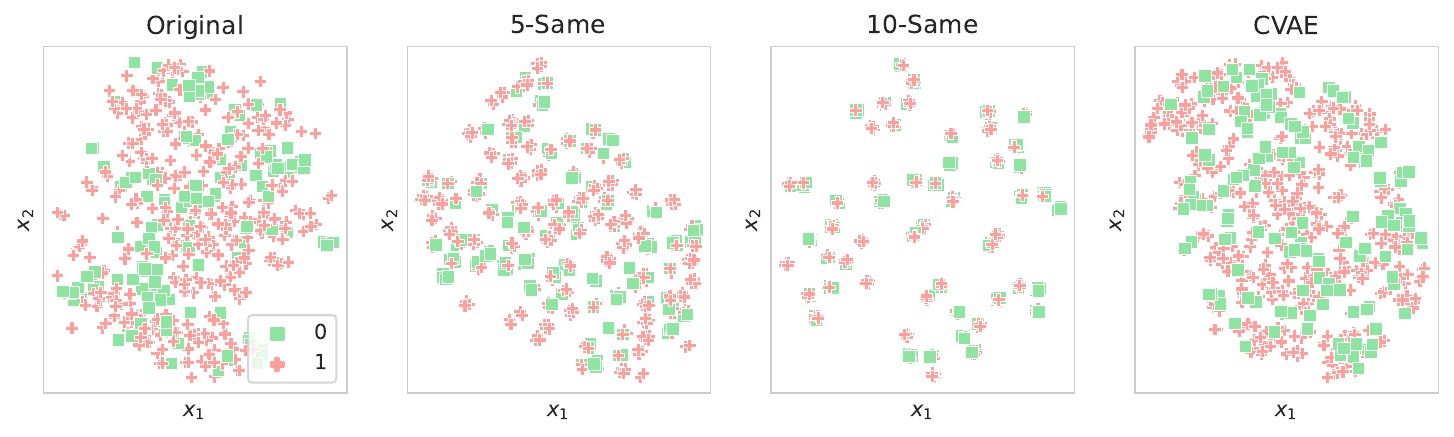}
	\caption{\label{fig:tsne} Class distribution of the \textit{BreastMNIST} dataset and its anonymous counterparts through $k$-Same $(5,10)$ and \texttt{CVAE}. Clearly, while \texttt{CVAE} faithfully preserves data diversity, $k$-Same tends to agglomerate information, increasing data sparsity and losing precious information, especially on limited-size datasets.}
\end{figure}

In addition to quantitative metrics, we conduct a qualitative analysis of the feature space, illustrated in Figure \ref{fig:tsne}, using t-distributed stochastic neighbor embedding (t-SNE) \cite{van2008visualizing}. Notably, our feature extractor was not originally trained on medical data, therefore it may encounter difficulties with tight class separation due to the subtle nature of diagnostic features. For instance, in the BreastMNIST dataset, which consists of breast ultrasound images, the fine-grained details necessary for accurate diagnosis are not always clearly represented. Nevertheless, subtle differences within the class still seem to be detectable, as indicated by the shape of small clusters scattered throughout the 2-dimensional feature space. 
With respect to the evaluated anonymization techniques, the limitations of the $k$-Same method, such as information loss and data sparsity, become particularly evident. Although DINOv2 ViT-B/14 may not well separate classes in such datasets, our \texttt{CVAE} approach manages to maintain the original distribution. Clearly, unlike $k$-Same, our method avoids information loss, ensuring our privacy-aware feature distribution closely mirrors the initial one.

\section{Discussion and conclusion} 

\textbf{Limitations} \, While our method does not directly expose training data, we did not rigorously investigate formal privacy guarantees. Furthermore, the effectiveness of \texttt{CVAE}s heavily depends on the capabilities of the chosen feature extractor. In addition, this approach could further benefit from domain-specific foundation models, as they may be able to capture more nuanced patterns. Lastly, our data-sharing framework assumes that all parties involved use the same feature extractor. However, with the increasing interest over \textit{general foundation models} \cite{tu2024towards}, this assumption becomes more reasonable and can be justified by the considerable benefits and enhanced downstream performance. \newline

\noindent \textbf{Conclusion} \, We demonstrate that conditional generative models, such as \texttt{CVAE}s, effectively address challenges related to privacy and data utility across a diverse range of datasets. This approach further shows high robustness even with small sample sizes and severe class imbalances. Future research can build upon our insights, investigating the sensitive role of the sampling variance, exploring further conditional generative models, or extending our approach to be trainable in an end-to-end manner for downstream tasks.

\section*{Acknowledgments}

This study was funded through the Hightech Agenda Bayern (HTA) of the Free State of Bavaria, Germany.


\bibliography{egbib}

\begin{thebibliography}{48}
\providecommand{\natexlab}[1]{#1}
\providecommand{\url}[1]{\texttt{#1}}
\expandafter\ifx\csname urlstyle\endcsname\relax
  \providecommand{\doi}[1]{doi: #1}\else
  \providecommand{\doi}{doi: \begingroup \urlstyle{rm}\Url}\fi

\bibitem[Altaf et~al.(2019)Altaf, Islam, Akhtar, and Janjua]{altaf2019going}
Fouzia Altaf, Syed~MS Islam, Naveed Akhtar, and Naeem~Khalid Janjua.
\newblock Going deep in medical image analysis: concepts, methods, challenges, and future directions.
\newblock \emph{IEEE Access}, 7:\penalty0 99540--99572, 2019.

\bibitem[Caron et~al.(2021)Caron, Touvron, Misra, J{\'e}gou, Mairal, Bojanowski, and Joulin]{caron2021emerging}
Mathilde Caron, Hugo Touvron, Ishan Misra, Herv{\'e} J{\'e}gou, Julien Mairal, Piotr Bojanowski, and Armand Joulin.
\newblock Emerging properties in self-supervised vision transformers.
\newblock In \emph{Proceedings of the IEEE/CVF international conference on computer vision}, pages 9650--9660, 2021.

\bibitem[Cimpoi et~al.(2014)Cimpoi, Maji, Kokkinos, Mohamed, and Vedaldi]{cimpoi14describing}
Mircea Cimpoi, Subhransu Maji, Iasonas Kokkinos, Sammy Mohamed, and Andrea Vedaldi.
\newblock Describing textures in the wild.
\newblock In \emph{Proceedings of the {IEEE} Conference on Computer Vision and Pattern Recognition ({CVPR})}, 2014.

\bibitem[Coates et~al.(2011)Coates, Ng, and Lee]{coates2011analysis}
Adam Coates, Andrew Ng, and Honglak Lee.
\newblock An analysis of single-layer networks in unsupervised feature learning.
\newblock In \emph{Proceedings of the fourteenth international conference on artificial intelligence and statistics}, pages 215--223. JMLR Workshop and Conference Proceedings, 2011.

\bibitem[Creswell et~al.(2018)Creswell, White, Dumoulin, Arulkumaran, Sengupta, and Bharath]{creswell2018generative}
Antonia Creswell, Tom White, Vincent Dumoulin, Kai Arulkumaran, Biswa Sengupta, and Anil~A Bharath.
\newblock Generative adversarial networks: An overview.
\newblock \emph{IEEE signal processing magazine}, 35\penalty0 (1):\penalty0 53--65, 2018.

\bibitem[Daneshjou et~al.(2022)Daneshjou, Vodrahalli, Novoa, Jenkins, Liang, Rotemberg, Ko, Swetter, Bailey, Gevaert, et~al.]{daneshjou2022disparities}
Roxana Daneshjou, Kailas Vodrahalli, Roberto~A Novoa, Melissa Jenkins, Weixin Liang, Veronica Rotemberg, Justin Ko, Susan~M Swetter, Elizabeth~E Bailey, Olivier Gevaert, et~al.
\newblock Disparities in dermatology ai performance on a diverse, curated clinical image set.
\newblock \emph{Science advances}, 8\penalty0 (31):\penalty0 eabq6147, 2022.

\bibitem[Doerrich et~al.(2024{\natexlab{a}})Doerrich, Archut, Di~Salvo, and Ledig]{doerricharchut2024kNNIntegration}
Sebastian Doerrich, Tobias Archut, Francesco Di~Salvo, and Christian Ledig.
\newblock Integrating knn with foundation models for adaptable and privacy-aware image classification.
\newblock In \emph{2024 IEEE 21th International Symposium on Biomedical Imaging (ISBI)}, 2024{\natexlab{a}}.

\bibitem[Doerrich et~al.(2024{\natexlab{b}})Doerrich, Di~Salvo, Brockmann, and Ledig]{doerrich2024rethinking}
Sebastian Doerrich, Francesco Di~Salvo, Julius Brockmann, and Christian Ledig.
\newblock Rethinking model prototyping through the medmnist+ dataset collection.
\newblock \emph{arXiv preprint arXiv:2404.15786}, 2024{\natexlab{b}}.

\bibitem[Dosovitskiy et~al.(2021)Dosovitskiy, Beyer, Kolesnikov, Weissenborn, Zhai, Unterthiner, Dehghani, Minderer, Heigold, Gelly, Uszkoreit, and Houlsby]{dosovitskiy2020image}
Alexey Dosovitskiy, Lucas Beyer, Alexander Kolesnikov, Dirk Weissenborn, Xiaohua Zhai, Thomas Unterthiner, Mostafa Dehghani, Matthias Minderer, Georg Heigold, Sylvain Gelly, Jakob Uszkoreit, and Neil Houlsby.
\newblock An image is worth 16x16 words: Transformers for image recognition at scale.
\newblock In \emph{International Conference on Learning Representations}, 2021.

\bibitem[Fajardo et~al.(2021)Fajardo, Findlay, Jaiswal, Yin, Houmanfar, Xie, Liang, She, and Emerson]{fajardo2021oversampling}
Val~Andrei Fajardo, David Findlay, Charu Jaiswal, Xinshang Yin, Roshanak Houmanfar, Honglei Xie, Jiaxi Liang, Xichen She, and David~B Emerson.
\newblock On oversampling imbalanced data with deep conditional generative models.
\newblock \emph{Expert Systems with Applications}, 169:\penalty0 114463, 2021.

\bibitem[Gkoulalas-Divanis et~al.(2014)Gkoulalas-Divanis, Loukides, and Sun]{gkoulalas2014publishing}
Aris Gkoulalas-Divanis, Grigorios Loukides, and Jimeng Sun.
\newblock Publishing data from electronic health records while preserving privacy: A survey of algorithms.
\newblock \emph{Journal of biomedical informatics}, 50:\penalty0 4--19, 2014.

\bibitem[Goodfellow et~al.(2020)Goodfellow, Pouget-Abadie, Mirza, Xu, Warde-Farley, Ozair, Courville, and Bengio]{goodfellow2020generative}
Ian Goodfellow, Jean Pouget-Abadie, Mehdi Mirza, Bing Xu, David Warde-Farley, Sherjil Ozair, Aaron Courville, and Yoshua Bengio.
\newblock Generative adversarial networks.
\newblock \emph{Communications of the ACM}, 63\penalty0 (11):\penalty0 139--144, 2020.

\bibitem[Gosselin et~al.(2022)Gosselin, Vieu, Loukil, and Benoit]{app12199901}
Rémi Gosselin, Loïc Vieu, Faiza Loukil, and Alexandre Benoit.
\newblock Privacy and security in federated learning: A survey.
\newblock \emph{Applied Sciences}, 12\penalty0 (19), 2022.
\newblock ISSN 2076-3417.

\bibitem[Gowal et~al.(2021)Gowal, Rebuffi, Wiles, Stimberg, Calian, and Mann]{gowal2021improving}
Sven Gowal, Sylvestre-Alvise Rebuffi, Olivia Wiles, Florian Stimberg, Dan~Andrei Calian, and Timothy~A Mann.
\newblock Improving robustness using generated data.
\newblock \emph{Advances in Neural Information Processing Systems}, 34:\penalty0 4218--4233, 2021.

\bibitem[Habib et~al.(2019)Habib, Karmakar, and Yearwood]{8758818}
Ahsan Habib, Chandan Karmakar, and John Yearwood.
\newblock Impact of ecg dataset diversity on generalization of cnn model for detecting qrs complex.
\newblock \emph{IEEE Access}, 7:\penalty0 93275--93285, 2019.

\bibitem[Hajihassnai et~al.(2021)Hajihassnai, Ardakanian, and Khazaei]{hajihassnai2021obscurenet}
Omid Hajihassnai, Omid Ardakanian, and Hamzeh Khazaei.
\newblock Obscurenet: Learning attribute-invariant latent representation for anonymizing sensor data.
\newblock In \emph{Proceedings of the International Conference on Internet-of-Things Design and Implementation}, IoTDI '21, pages 40--52, New York, NY, USA, 2021. Association for Computing Machinery.
\newblock ISBN 9781450383547.

\bibitem[Hendrycks and Dietterich(2019)]{hendrycks2019benchmarking}
Dan Hendrycks and Thomas Dietterich.
\newblock Benchmarking neural network robustness to common corruptions and perturbations.
\newblock \emph{Proceedings of the International Conference on Learning Representations}, 2019.

\bibitem[Jeon et~al.(2022)Jeon, Park, Kim, Morley, and Cho]{jeon2022k}
Minkyu Jeon, Hyeonjin Park, Hyunwoo~J Kim, Michael Morley, and Hyunghoon Cho.
\newblock k-salsa: k-anonymous synthetic averaging of retinal images via local style alignment.
\newblock In \emph{European Conference on Computer Vision}, pages 661--678. Springer, 2022.

\bibitem[Kim and Yang(2019)]{8744221}
Taehoon Kim and Jihoon Yang.
\newblock Latent-space-level image anonymization with adversarial protector networks.
\newblock \emph{IEEE Access}, 7:\penalty0 84992--84999, 2019.

\bibitem[Larson et~al.(2020)Larson, Zheng, Mahendran, Tekriwal, Cheung, Guldan, Leach, and Kummerfeld]{larson-etal-2020-iterative}
Stefan Larson, Anthony Zheng, Anish Mahendran, Rishi Tekriwal, Adrian Cheung, Eric Guldan, Kevin Leach, and Jonathan~K. Kummerfeld.
\newblock Iterative feature mining for constraint-based data collection to increase data diversity and model robustness.
\newblock In \emph{Proceedings of the 2020 Conference on Empirical Methods in Natural Language Processing (EMNLP)}, pages 8097--8106, Online, November 2020. Association for Computational Linguistics.

\bibitem[Li et~al.(2024)Li, Wong, Zhang, Usuyama, Liu, Yang, Naumann, Poon, and Gao]{li2024llava}
Chunyuan Li, Cliff Wong, Sheng Zhang, Naoto Usuyama, Haotian Liu, Jianwei Yang, Tristan Naumann, Hoifung Poon, and Jianfeng Gao.
\newblock Llava-med: Training a large language-and-vision assistant for biomedicine in one day.
\newblock \emph{Advances in Neural Information Processing Systems}, 36, 2024.

\bibitem[Mai et~al.(2022)Mai, Cundy, Choi, Hu, Lao, and Ermon]{mai2022towards}
Gengchen Mai, Chris Cundy, Kristy Choi, Yingjie Hu, Ni~Lao, and Stefano Ermon.
\newblock Towards a foundation model for geospatial artificial intelligence (vision paper).
\newblock In \emph{Proceedings of the 30th International Conference on Advances in Geographic Information Systems}, pages 1--4, 2022.

\bibitem[Maji et~al.(2013)Maji, Kannala, Rahtu, Blaschko, and Vedaldi]{maji13fine-grained}
S.~Maji, J.~Kannala, E.~Rahtu, M.~Blaschko, and A.~Vedaldi.
\newblock Fine-grained visual classification of aircraft.
\newblock Technical report, 2013.

\bibitem[Meden et~al.(2018)Meden, Emer{\v{s}}i{\v{c}}, {\v{S}}truc, and Peer]{meden2018k}
Bla{\v{z}} Meden, {\v{Z}}iga Emer{\v{s}}i{\v{c}}, Vitomir {\v{S}}truc, and Peter Peer.
\newblock k-same-net: k-anonymity with generative deep neural networks for face deidentification.
\newblock \emph{Entropy}, 20\penalty0 (1):\penalty0 60, 2018.

\bibitem[Nakata et~al.(2022)Nakata, Ng, Miyashita, Maki, Lin, and Deguchi]{nakata2022revisiting}
Kengo Nakata, Youyang Ng, Daisuke Miyashita, Asuka Maki, Yu-Chieh Lin, and Jun Deguchi.
\newblock Revisiting a knn-based image classification system with high-capacity storage.
\newblock In \emph{European Conference on Computer Vision}, pages 457--474. Springer, 2022.

\bibitem[Nasr et~al.(2019)Nasr, Shokri, and Houmansadr]{nasr2019comprehensive}
Milad Nasr, Reza Shokri, and Amir Houmansadr.
\newblock Comprehensive privacy analysis of deep learning: Passive and active white-box inference attacks against centralized and federated learning.
\newblock In \emph{2019 IEEE symposium on security and privacy (SP)}, pages 739--753. IEEE, 2019.

\bibitem[Newton et~al.(2005)Newton, Sweeney, and Malin]{newton2005preserving}
Elaine~M Newton, Latanya Sweeney, and Bradley Malin.
\newblock Preserving privacy by de-identifying face images.
\newblock \emph{IEEE transactions on Knowledge and Data Engineering}, 17\penalty0 (2):\penalty0 232--243, 2005.

\bibitem[Nguyen et~al.(2023)Nguyen, Brandstetter, Kapoor, Gupta, and Grover]{nguyen2023climax}
Tung Nguyen, Johannes Brandstetter, Ashish Kapoor, Jayesh~K Gupta, and Aditya Grover.
\newblock Climax: A foundation model for weather and climate.
\newblock \emph{arXiv preprint arXiv:2301.10343}, 2023.

\bibitem[Oquab et~al.(2023)Oquab, Darcet, Moutakanni, Vo, Szafraniec, Khalidov, Fernandez, Haziza, Massa, El-Nouby, et~al.]{oquab2023dinov2}
Maxime Oquab, Timoth{\'e}e Darcet, Th{\'e}o Moutakanni, Huy Vo, Marc Szafraniec, Vasil Khalidov, Pierre Fernandez, Daniel Haziza, Francisco Massa, Alaaeldin El-Nouby, et~al.
\newblock Dinov2: Learning robust visual features without supervision.
\newblock \emph{arXiv preprint arXiv:2304.07193}, 2023.

\bibitem[Parkhi et~al.(2012)Parkhi, Vedaldi, Zisserman, and Jawahar]{parkhi2012cats}
Omkar~M Parkhi, Andrea Vedaldi, Andrew Zisserman, and CV~Jawahar.
\newblock Cats and dogs.
\newblock In \emph{2012 IEEE conference on computer vision and pattern recognition}, pages 3498--3505. IEEE, 2012.

\bibitem[Paul and Chen(2022)]{paul2022vision}
Sayak Paul and Pin-Yu Chen.
\newblock Vision transformers are robust learners.
\newblock In \emph{Proceedings of the AAAI conference on Artificial Intelligence}, volume~36, pages 2071--2081, 2022.

\bibitem[Pennisi et~al.(2023)Pennisi, Proietto~Salanitri, Bellitto, Palazzo, Bagci, and Spampinato]{pennisi2023privacy}
Matteo Pennisi, Federica Proietto~Salanitri, Giovanni Bellitto, Simone Palazzo, Ulas Bagci, and Concetto Spampinato.
\newblock A privacy-preserving walk in the latent space of generative models for medical applications.
\newblock In \emph{International Conference on Medical Image Computing and Computer-Assisted Intervention}, pages 422--431. Springer, 2023.

\bibitem[Perero-Codosero et~al.(2022)Perero-Codosero, Espinoza-Cuadros, and Hern{\'a}ndez-G{\'o}mez]{perero2022x}
Juan~M Perero-Codosero, Fernando~M Espinoza-Cuadros, and Luis~A Hern{\'a}ndez-G{\'o}mez.
\newblock X-vector anonymization using autoencoders and adversarial training for preserving speech privacy.
\newblock \emph{Computer Speech \& Language}, 74:\penalty0 101351, 2022.

\bibitem[Pesteie et~al.(2019)Pesteie, Abolmaesumi, and Rohling]{pesteie2019adaptive}
Mehran Pesteie, Purang Abolmaesumi, and Robert~N Rohling.
\newblock Adaptive augmentation of medical data using independently conditional variational auto-encoders.
\newblock \emph{IEEE transactions on medical imaging}, 38\penalty0 (12):\penalty0 2807--2820, 2019.

\bibitem[Radford et~al.(2021)Radford, Kim, Hallacy, Ramesh, Goh, Agarwal, Sastry, Askell, Mishkin, Clark, et~al.]{radford2021learning}
Alec Radford, Jong~Wook Kim, Chris Hallacy, Aditya Ramesh, Gabriel Goh, Sandhini Agarwal, Girish Sastry, Amanda Askell, Pamela Mishkin, Jack Clark, et~al.
\newblock Learning transferable visual models from natural language supervision.
\newblock In \emph{International conference on machine learning}, pages 8748--8763. PMLR, 2021.

\bibitem[Rodr{\'\i}guez-Barroso et~al.(2023)Rodr{\'\i}guez-Barroso, Jim{\'e}nez-L{\'o}pez, Luz{\'o}n, Herrera, and Mart{\'\i}nez-C{\'a}mara]{rodriguez2023survey}
Nuria Rodr{\'\i}guez-Barroso, Daniel Jim{\'e}nez-L{\'o}pez, M~Victoria Luz{\'o}n, Francisco Herrera, and Eugenio Mart{\'\i}nez-C{\'a}mara.
\newblock Survey on federated learning threats: Concepts, taxonomy on attacks and defences, experimental study and challenges.
\newblock \emph{Information Fusion}, 90:\penalty0 148--173, 2023.

\bibitem[Saxena and Cao(2021)]{saxena2021generative}
Divya Saxena and Jiannong Cao.
\newblock Generative adversarial networks (gans) challenges, solutions, and future directions.
\newblock \emph{ACM Computing Surveys (CSUR)}, 54\penalty0 (3):\penalty0 1--42, 2021.

\bibitem[Sweeney(2002)]{sweeney2002k}
Latanya Sweeney.
\newblock k-anonymity: A model for protecting privacy.
\newblock \emph{International journal of uncertainty, fuzziness and knowledge-based systems}, 10\penalty0 (05):\penalty0 557--570, 2002.

\bibitem[Thanh-Tung and Tran(2020)]{9207181}
Hoang Thanh-Tung and Truyen Tran.
\newblock Catastrophic forgetting and mode collapse in gans.
\newblock In \emph{2020 International Joint Conference on Neural Networks (IJCNN)}, 2020.

\bibitem[Tu et~al.(2024)Tu, Azizi, Driess, Schaekermann, Amin, Chang, Carroll, Lau, Tanno, Ktena, et~al.]{tu2024towards}
Tao Tu, Shekoofeh Azizi, Danny Driess, Mike Schaekermann, Mohamed Amin, Pi-Chuan Chang, Andrew Carroll, Charles Lau, Ryutaro Tanno, Ira Ktena, et~al.
\newblock Towards generalist biomedical ai.
\newblock \emph{NEJM AI}, 1\penalty0 (3):\penalty0 AIoa2300138, 2024.

\bibitem[Van~der Maaten and Hinton(2008)]{van2008visualizing}
Laurens Van~der Maaten and Geoffrey Hinton.
\newblock Visualizing data using t-sne.
\newblock \emph{Journal of machine learning research}, 9\penalty0 (11), 2008.

\bibitem[Woerner et~al.(2024)Woerner, Jaques, and Baumgartner]{woerner2024comprehensive}
Stefano Woerner, Arthur Jaques, and Christian~F Baumgartner.
\newblock A comprehensive and easy-to-use multi-domain multi-task medical imaging meta-dataset (medimeta).
\newblock \emph{arXiv preprint arXiv:2404.16000}, 2024.

\bibitem[Yang et~al.(2023)Yang, Shi, Wei, Liu, Zhao, Ke, Pfister, and Ni]{yang2023medmnist}
Jiancheng Yang, Rui Shi, Donglai Wei, Zequan Liu, Lin Zhao, Bilian Ke, Hanspeter Pfister, and Bingbing Ni.
\newblock Medmnist v2-a large-scale lightweight benchmark for 2d and 3d biomedical image classification.
\newblock \emph{Scientific Data}, 10\penalty0 (1):\penalty0 41, 2023.

\bibitem[Yang et~al.(2022)Yang, Niu, Qiu, Song, Zhang, Tian, and Guo]{yang2022cluster}
Yuanzhe Yang, Zhiyi Niu, Yuying Qiu, Biao Song, Xinchang Zhang, Yuan Tian, and Ran Guo.
\newblock A cluster-based facial image anonymization method using variational autoencoder.
\newblock In \emph{International Conference on Big Data and Security}, pages 621--633. Springer, 2022.

\bibitem[Yu et~al.(2022)Yu, Khadivi, and Xu]{yu-etal-2022-data}
Yu~Yu, Shahram Khadivi, and Jia Xu.
\newblock Can data diversity enhance learning generalization?
\newblock In \emph{Proceedings of the 29th International Conference on Computational Linguistics}, Gyeongju, Republic of Korea, October 2022. International Committee on Computational Linguistics.

\bibitem[Zhang et~al.(2021)Zhang, Xie, Bai, Yu, Li, and Gao]{zhang2021survey}
Chen Zhang, Yu~Xie, Hang Bai, Bin Yu, Weihong Li, and Yuan Gao.
\newblock A survey on federated learning.
\newblock \emph{Knowledge-Based Systems}, 216:\penalty0 106775, 2021.

\bibitem[Zhang et~al.(2018)Zhang, Isola, Efros, Shechtman, and Wang]{zhang2018perceptual}
Richard Zhang, Phillip Isola, Alexei~A Efros, Eli Shechtman, and Oliver Wang.
\newblock The unreasonable effectiveness of deep features as a perceptual metric.
\newblock In \emph{CVPR}, 2018.

\bibitem[Zhou et~al.(2023)Zhou, Chia, Wagner, Ayhan, Williamson, Struyven, Liu, Xu, Lozano, Woodward-Court, et~al.]{zhou2023foundation}
Yukun Zhou, Mark~A Chia, Siegfried~K Wagner, Murat~S Ayhan, Dominic~J Williamson, Robbert~R Struyven, Timing Liu, Moucheng Xu, Mateo~G Lozano, Peter Woodward-Court, et~al.
\newblock A foundation model for generalizable disease detection from retinal images.
\newblock \emph{Nature}, 622\penalty0 (7981):\penalty0 156--163, 2023.

\end{thebibliography}

\end{document}